\documentclass[10pt,journal,compsoc]{IEEEtran}

\ifCLASSOPTIONcompsoc
  \usepackage[nocompress]{cite}
\else
  \usepackage{cite}
\fi

\ifCLASSINFOpdf

\else

\fi

\usepackage{epsfig}
\usepackage{graphicx}
\usepackage{amsmath}
\usepackage{amssymb}
\usepackage{comment}
\usepackage{multirow}

\usepackage{booktabs}
\usepackage{array, caption, threeparttable}
\usepackage{caption}

\usepackage{algorithm}
\usepackage{listings}
\usepackage{color}

\usepackage{mathtools}

\usepackage{todonotes}
\usepackage{gensymb}

\begin{document}

\title{Revisiting Pixel-Wise Supervision for Face Anti-Spoofing}

\author{Zitong Yu,~\IEEEmembership{Student Member,~IEEE}, Xiaobai Li,~\IEEEmembership{Member,~IEEE}, Jingang Shi, \\ Zhaoqiang Xia,~\IEEEmembership{Member,~IEEE} and Guoying Zhao,~\IEEEmembership{Senior Member,~IEEE}


\IEEEcompsocitemizethanks{\IEEEcompsocthanksitem Z. Yu, X. Li and G. Zhao are with Center for Machine Vision and Signal Analysis, University of Oulu, Oulu 90014, Finland.

E-mail: \{zitong.yu, xiaobai.li,  guoying.zhao\}@oulu.fi.

\IEEEcompsocthanksitem J. Shi is with the School of Software Engineering, Xi'an Jiaotong University, Xi'an 710049, China. E-mail: jingang@xjtu.edu.cn.

\IEEEcompsocthanksitem Z. Xia is with the School of Electronics and Information, Northwestern Polytechnical University, Xi'an, China. E-mail: zxia@nwpu.edu.cn.

}
\thanks{Manuscript received November 25, 2020.\\ (Corresponding author: Guoying Zhao)}}

\markboth{IEEE Transactions on ,~Vol.~*, No.~*, **}%
{Shell \MakeLowercase{\textit{et al.}}: Bare Advanced Demo of IEEEtran.cls for IEEE Computer Society Journals}

\IEEEtitleabstractindextext{%
\begin{abstract}

Face anti-spoofing (FAS) plays a vital role in securing face recognition systems from the presentation attacks (PAs). As more and more realistic PAs with novel types spring up, it is necessary to develop robust algorithms for detecting unknown attacks even in unseen scenarios. However, deep models supervised by traditional binary loss (e.g., `0' for bonafide vs. `1' for PAs) are weak in describing intrinsic and discriminative spoofing patterns. Recently, pixel-wise supervision has been proposed for the FAS task, intending to provide more fine-grained pixel/patch-level cues. In this paper, we firstly give a comprehensive review and analysis about the existing pixel-wise supervision methods for FAS. Then we propose a novel pyramid supervision, which guides deep models to learn both local details and global semantics from multi-scale spatial context. Extensive experiments are performed on five FAS benchmark datasets to show that, without bells and whistles, the proposed pyramid supervision could not only improve the performance beyond existing pixel-wise supervision frameworks, but also enhance the model's interpretability (i.e., locating the patch-level positions of PAs more reasonably). Furthermore, elaborate studies are conducted for exploring the efficacy of different architecture configurations with two kinds of pixel-wise supervisions (binary mask and depth map supervisions), which provides inspirable insights for future architecture/supervision design.

\end{abstract}

\begin{IEEEkeywords}
face anti-spoofing, pixel-wise supervision, pyramid supervision.
\end{IEEEkeywords}}

\maketitle

\IEEEdisplaynontitleabstractindextext

%
\IEEEpeerreviewmaketitle

\ifCLASSOPTIONcompsoc
\IEEEraisesectionheading{\section{Introduction}\label{sec:introduction}}
\else
\section{Introduction}
\label{sec:introduction}
\fi

\IEEEPARstart{D}{ue} to its convenience and remarkable accuracy, face recognition technology~\cite{guo2020learning} has applied in a few interactive intelligent applications such as checking-in and mobile payment. However, most existing face recognition systems are vulnerable to presentation attacks (PAs) ranging from print, replay, makeup and 3D-mask attacks~\cite{jia2020survey}. Therefore, both the academia and industry have paid attention on the development of face anti-spoofing (FAS) technology for securing the face recognition system.

In the past two decades, both traditional~\cite{pan2007eye,li2016generalized,Pereira2012LBP,Komulainen2014Context,Patel2016Secure} and deep learning-based~\cite{yu2020searching,yu2020face,Liu2018Learning,yang2019face,Atoum2018Face,yu2020multi,zhang2020casia} methods have shown their effectiveness for presentation attack detection (PAD). Most traditional algorithms focus on human liveness cues and handcrafted features, which need rich task-aware prior knowledge. In term of the methods based on the liveness cues, eye-blinking~\cite{pan2007eye,jee2006liveness,li2008eye}, face and head movement~\cite{wang2009face,bao2009liveness} (e.g., nodding and smiling), gaze tracking~\cite{bigun2004assuring,ali2012liveness} and remote physiological signals (e.g., rPPG~\cite{li2016generalized,Liu2018Learning,lin2019face,yu2019remote1}) are explored as dynamic discrimination. However, these physiological liveness cues are usually captured from long-term interactive face videos and easily mimicked from the video attacks, making them less reliable and inconvenient for practical deployment. On the other hand, classical handcrafted descriptors (e.g., LBP~\cite{boulkenafet2015face,Pereira2012LBP}, 
SIFT~\cite{Patel2016Secure},  HOG~\cite{Komulainen2014Context} and DoG~\cite{tan2010face}) are designed for extracting effective spoofing patterns from various color spaces (RGB, HSV, and YCbCr). Although such handcrafted features could be cascaded with a trained classifier (e.g., SVM~\cite{suykens1999least}) efficiently, they still suffer from limited representation capacity and are vulnerable under unseen scenarios and unknown PAs.  

\begin{figure}
\centering

\includegraphics[scale=0.35]{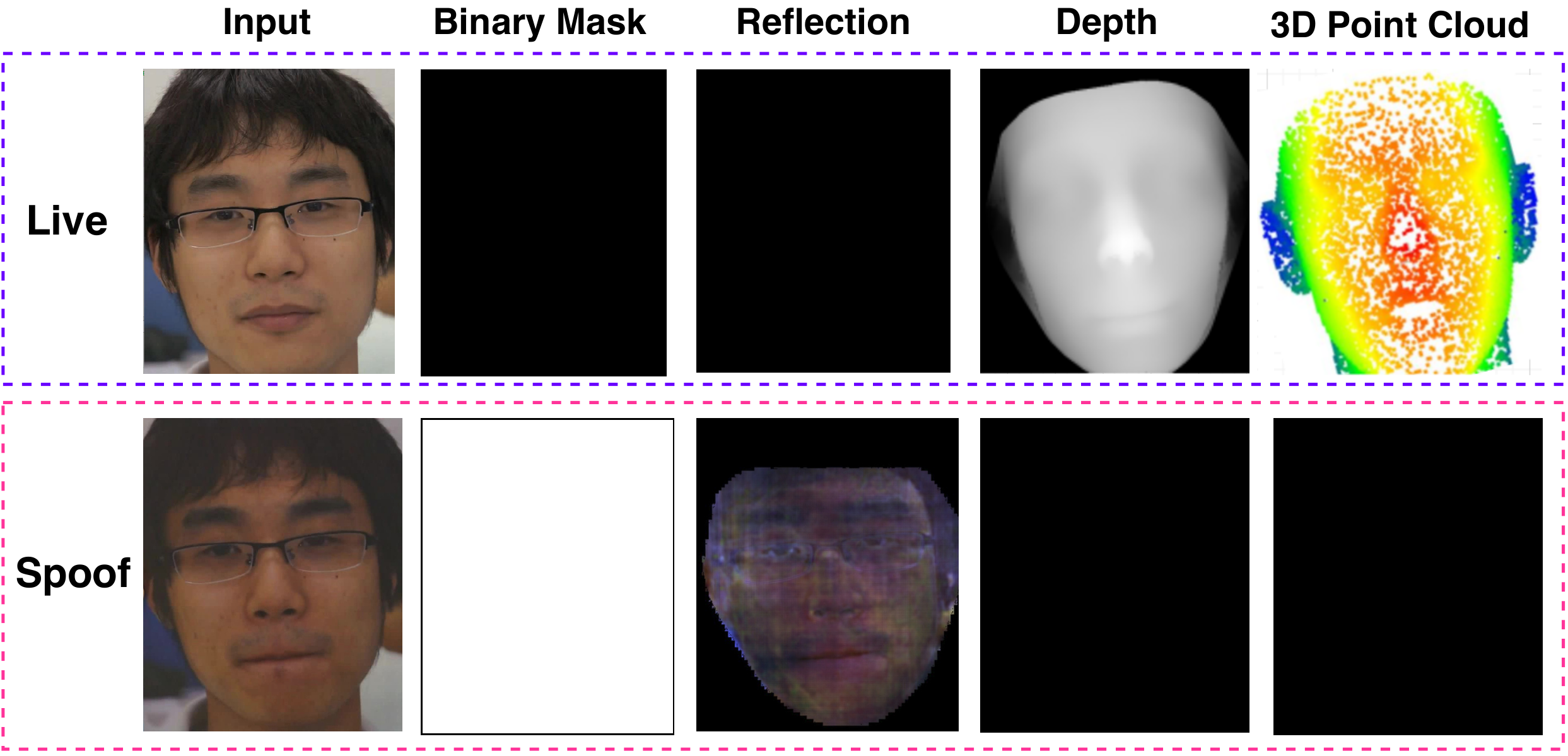}
  \caption{ 
  Visualization of different kinds of pixel-wise supervision for FAS. Given an RGB live/spoof face image as input, the deep models could be supervised by  binary mask label~\cite{george2019deep,liu2019deep,sun2020face}, pseudo reflection maps~\cite{yu2020face,kim2019basn}, pseudo depth labels~\cite{Atoum2018Face,Liu2018Learning} and 3D point cloud maps~\cite{li3dpc}. 
  }
\label{fig:Figure1}
\end{figure}

In contrast, a few deep learning based methods are proposed for both frame and video level face liveness detection. Most works\cite{yang2014learn,Li2017An,Patel2016Cross,george2019deep,jourabloo2018face,jia20203d,li2020compactnet} treat FAS as a binary classification problem (e.g., `0' for live while `1' for spoofing faces) thus supervised by a simple binary cross-entropy loss. However, convolutional neural networks (CNNs) with binary loss might discover arbitrary cues that are able to separate the two classes (e.g., screen bezel), but not the faithful spoofing patterns. Recently, pixel-wise supervision~\cite{Atoum2018Face,Liu2018Learning,yu2020face,kim2019basn,george2019deep,yu2020fas2} attracts more attention as it provides more fine-grained context-aware supervision signals, which is beneficial for deep models learning intrinsic spoofing cues. As can be seen from Fig.~\ref{fig:Figure1}, pseudo depth labels~\cite{Atoum2018Face,Liu2018Learning}, reflection maps~\cite{yu2020face,kim2019basn}, binary mask label~\cite{george2019deep,liu2019deep,sun2020face} and 3D point cloud maps~\cite{li3dpc} are typical pixel-wise supervision signals, which describes the local live/spoofing cues in pixel/patch level. Despite with gained performance, existing pixel-wise supervisions still have the representation gaps between local context and global semantics, and are easily influenced by the local disturbance.

For the perspective of evaluation protocols, most existing FAS methods are supervised by predefined scenarios and PAs. Thus, the trained models probabilistically overfit on several specific domains and attacks, which are vulnerable to domain shift and unseen attacks. As FAS is actually an open-set problem in practice, the uncertain gaps (e.g., environments and attack types) between training and testing conditions should be considered. On one hand, adversarial learning~\cite{shao2019multi}, fine-grained meta learning~\cite{shao2019regularized} and  multi-domain disentangled learning~\cite{wang2020cross} are utilized to learn robust features for domain generalization in FAS. On the other hand, in order to detect unseen attacks successfully, one class SVM~\cite{arashloo2017anomaly}, deep tree network~\cite{liu2019deep} and adaptive inner-update meta learning~\cite{qin2019learning} are developed. Despite enhancing the generalization capacity via learning strategies~\cite{shao2019multi,shao2019regularized,qin2019learning,wang2020cross}, they are still hard to explicitly learn detailed intrinsic spoofing patterns with coarse supervision signals. Moreover, it is not trivial to explore and understand how the pixel-wise supervision impacts the mainstream architectures on various FAS evaluation protocols.

Motivated by the discussions above, we firstly investigate existing pixel-wise supervision methods for FAS, which aims to help FAS researchers conveniently compare and understand pixel-wise supervision based state of the arts. Furthermore, we propose a novel pyramid supervision, which is able to plug-and-play into the current pixel-wise supervision framework flexibly and provide multi-scale patch/global signals. To sum up, the main contributions of this paper are listed:

\begin{itemize}
    
    \item We give a comprehensive review of the existing pixel-wise supervision methods in the FAS task.
    
    \item  We propose a novel plugged-and-played pyramid supervision, which is able to provide richer multi-scale spatial context for fine-grained learning. The proposed pyramid supervision could not only improve the performance beyond the existing pixel-wise supervision framework, but also enhance the model's interpretability.
    
   \item  We conduct elaborate studies for exploring the efficacy of different architecture configurations with pixel-wise supervisions, which provides the insights for future architecture/supervision design.
 
\end{itemize}

In the rest of the paper, Section~\ref{sec:relatedwork} provides a review of pixel-wise supervision for FAS. Section~\ref{sec:pyramid} introduces the pyramid supervision for multiple kinds of pixel-wise supervisions (e.g., binary mask maps and pseudo depth maps). Section~\ref{sec:experiment} provides rigorous ablation studies and evaluates the performance of the proposed pyramid supervision on five benchmark datasets. Finally, conclusions and future works are given in Section~\ref{sec:conclusion}.

\section{review of pixel-wise supervision} \label{sec:relatedwork}
In this section, we will review existing FAS methods with pixel-wise supervision including depth map, binary mask and so on. The relevant summary is shown in Table~\ref{tab:review}.

\noindent\textbf{Supervised by Depth Map.}\quad  
According to the human prior knowledge of FAS, most PAs (e.g., plain printed paper and electronic screen) merely have no real facial depth information, which could be utilized as discriminative supervision signals. As a result, some recent works~\cite{Atoum2018Face,peng2020ts,yu2020searching,wang2020deep} adopt pixel-wise pseudo depth labels (forth column in Fig.~\ref{fig:Figure1}) to guide the deep models, enforcing them predict the real depth for live samples while zero maps for spoof ones. Atoum et al.~\cite{Atoum2018Face} first leveraged pseudo depth labels to guide the multi-scale fully convolutional network (in this paper we call it as `DepthNet' for simplicity). Thus, the well-trained DepthNet is able to predict holistic depth maps as decision evidence. Based on the generated pseudo depth labels, Wang et al.~\cite{wang2020deep} designed a contrastive depth loss (CDL) for exploiting fine-grained local depth cues, which also has been applied in training the central difference convolutional networks (CDCN)~\cite{yu2020searching}. Similarly, Peng et al.~\cite{peng2020ts} fused the depth supervised stream with another color (RGB,HSV, YCbCr) stream to obtain more robust representation. 

Despite with better interpretability and performance compared with traditional binary label, pixel-wise depth label still exists two issues: 1) synthesis of 3D shapes for every training sample is costly and not accurate enough; and 2) it lacks the reasonability for some PAs with real depth (e.g., 3D mask and Mannequin).

\newcommand{\tabincell}[2]{\begin{tabular}{@{}#1@{}}#2\end{tabular}}
\begin{table*}[t]
\centering
\caption{Summary of the representative face anti-spoofing methods with \textbf{pixel-wise supervision}. `S/D' is short for Static/Dynamic. `NAS' denotes neural searched architecture. Note that some methods also consider classification loss (e.g., binary cross entropy loss, triplet loss, and adversarial loss) which are not listed in the `Supervision' column.} \label{tab:review}
\resizebox{1.0\textwidth}{!} {\begin{tabular}{l c c c c c c} 
 \toprule
 Method & Year & Supervision & Backbone & Input & S/D & Description \\
 \midrule
 Depth\&Patch~\cite{Atoum2018Face} & 2017 & Depth & \tabincell{c}{PatchNet\\DepthNet} & \tabincell{c}{YCbCr\\HSV} & S & \tabincell{c}{ local patch features and holistic\\ depth maps extracted by two-stream CNNs}\\

 \midrule
 Auxiliary~\cite{Liu2018Learning} & 2018 & \tabincell{c}{Depth\\rPPG spectrum} & DepthNet & \tabincell{c}{RGB\\HSV} & D & \tabincell{c}{local temporal features learned from CNN-RNN model \\with pixel-wise depth and sequence-wise rPPG supervision}\\
 
  \midrule
 De-Spoof~\cite{jourabloo2018face} & 2018 & \tabincell{c}{Depth\\FourierMap} & \tabincell{c}{DSNet\\DepthNet} & \tabincell{c}{RGB\\HSV} & S & \tabincell{c}{subtle spoof noise estimated from CNN with proper supervisions}\\

  \midrule
 BASN~\cite{kim2019basn} & 2019 & \tabincell{c}{Depth\\Reflection} & \tabincell{c}{DepthNet\\Enrichment} & \tabincell{c}{RGB\\HSV} & S & \tabincell{c}{generalizable features via bipartite auxiliary supervision}\\

 \midrule
Reconstruction~\cite{chen2019towards} & 2019 &  \tabincell{c}{RGB Input (live)\\ZeroMap (spoof)}   & U-Net  & RGB & S & multi-level semantic features from autoencoder\\

  \midrule
DTN~\cite{liu2019deep} & 2019 & BinaryMask & Tree Network  & \tabincell{c}{RGB\\HSV} & S & \tabincell{c}{partition the spoof samples into semantic \\sub-groups in an unsupervised fashion}\\

  \midrule
 PixBiS~\cite{george2019deep} & 2019 & BinaryMask & DenseNet161  & RGB & S & \tabincell{c}{deep pixel-wise binary supervision without trivial depth synthesis}\\

 \midrule
 A-PixBiS~\cite{hossaindeeppixbis} & 2020 & BinaryMask & DenseNet161  & RGB & S & \tabincell{c}{incorporate a variant of binary cross entropy that  enforces\\ a margin in angular space for attentive pixel wise supervision}\\

 \midrule
Auto-FAS~\cite{yu2020auto2} & 2020 & BinaryMask  & NAS  & RGB & S & \tabincell{c}{well-suitable lightweight networks searched for mobile-level FAS}\\

 \midrule
MRCNN~\cite{ma2020novel} & 2020 &  BinaryMask  & Shallow CNN  & RGB & S & \tabincell{c}{introduces local losses to patches, and constraints the \\ entire face region to avoid over-emphasizing certain local areas}\\

 \midrule
STDN~\cite{liu2020disentangling} & 2020 &  BinaryMask   & \tabincell{c}{U-Net\\PatchGAN}  & RGB & S & disentangled spoof trace via adversarial learning\\

 \midrule
FCN-LSA~\cite{sun2020face2} & 2020 &  BinaryMask  & DepthNet  & RGB & S & high frequent spoof cues from  lossless size adaptation module\\

 \midrule
CDCN~\cite{yu2020searching} & 2020 &  Depth  & DepthNet  & RGB & S & \tabincell{c}{intrinsic detailed patterns via aggregating both intensity and\\ gradient information from stacked central difference convolutions. }\\

 \midrule
MM-CDCN~\cite{yu2020multi} & 2020 &  BinaryMask  & DepthNet  &  \tabincell{c}{RGB\\NIR\\Depth} & S & \tabincell{c}{ multimodal features from central difference convolutional networks }\\

\midrule
FAS-SGTD~\cite{wang2020deep} & 2020 &  Depth  &  \tabincell{c}{DepthNet\\STPM}  & RGB & D & \tabincell{c}{detailed discriminative dynamics cues from stacked Residual \\Spatial Gradient Block and Spatio-Temporal Propagation Module}\\

 \midrule
TS-FEN~\cite{peng2020ts} & 2020 &  Depth   & \tabincell{c}{ResNet34\\FCN}  & \tabincell{c}{RGB\\YCbCr\\HSV} & S & \tabincell{c}{discriminative fused features from \\depth-stream and chroma-stream networks}\\

 \midrule
SAPLC~\cite{sun2020face} & 2020 &  TernaryMap   & DepthNet  & \tabincell{c}{RGB\\HSV} & S & \tabincell{c}{accurate image-level decision via spatial aggregation of \\pixel-level local classifiers even with insufficient training samples}\\

 \midrule
BCN~\cite{yu2020face} & 2020 &  \tabincell{c}{BinaryMask\\Depth\\Reflection}   & DepthNet  & RGB & S & \tabincell{c}{intrinsic material-based patterns captured via \\aggregating
multi-level bilateral macro- and micro- information}\\

 \midrule
Disentangled~\cite{zhang2020face} & 2020 &  \tabincell{c}{Depth\\TextureMap}   & DepthNet  & RGB & S & \tabincell{c}{liveness and content features via disentangled representation learning}\\

 \midrule
AENet~\cite{zhang2020celeba} & 2020 &  \tabincell{c}{Depth\\Reflection}   & ResNet18  & RGB & S & \tabincell{c}{rich semantic features using Auxiliary Information,\\ Embedding Network with multi-task learning framework}\\

 \midrule
LGSC~\cite{feng2020learning} & 2020 &  ZeroMap (live)   & U-Net(ResNet18)  & RGB & S & \tabincell{c}{discriminative live-spoof differences learned within a residual-\\learning framework with the perspective of anomaly detection }\\

 \midrule
3DPC-Net~\cite{li3dpc} & 2020 &  3D Point Cloud   & ResNet18  & RGB & S & discriminative features via fine-grained 3D Point Cloud supervision\\

 \bottomrule
 \end{tabular}}
\end{table*}

\noindent\textbf{Supervised by Binary Mask.}\quad Compared with pseudo depth map, binary mask label~\cite{liu2019deep,george2019deep,hossaindeeppixbis,yu2020auto2,ma2020novel,liu2020disentangling,sun2020face2} is easier to be generated and more generalizable to all PAs. To be specific, the binary supervision would be provided for the deep embedding features in each spatial position (second column in Fig.~\ref{fig:Figure1}). In other words, through the binary mask label, we can find whether PAs occur in the corresponding patches, which is attack-type-agnostic and spatially interpretable. George and Marcel~\cite{george2019deep} were the first to introduce deep pixel-wise binary supervision (PixBis), which assisted to predict the intermediate confidence map for the cascaded final binary classification. Similarly, Liu et al.~\cite{liu2019deep} constrained the leaf nodes with both binary classification and pixel-wise mask regression. Hossaind et al.~\cite{hossaindeeppixbis} proposed to add an attention module for feature refinement before calculating the deep pixel-wise binary loss. Yu et al.~\cite{yu2020auto2} searched lightweight FAS architectures with pixel-wise binary supervision. Liu et al.~\cite{liu2020disentangling} utilized Early Spoof Regressor with pixel-wise binary supervision to enhance discriminativeness of the generator. Ma et al.~\cite{ma2020novel} proposed a multi-regional CNN with the local binary classification loss to local patches. Yu et al.~\cite{yu2020multi} utilized pixel-wise binary label to supervise the multimodal CDCN and won the first place in the ChaLearn multi-modal face anti-spoofing attack detection challenge @CVPR2020~\cite{liu2020cross}.

With the help of spatially positional knowledge, binary mask label not only boosts the models' discrimination, but also benefits neural architecture search. As the types of PAs could be defined as fine-grained multiple classes, it is also worth exploring whether extending binary maps to multi-class maps is beneficial for supervising the PAs detector.

\begin{figure*}[t]
\centering
\includegraphics[scale=0.4]{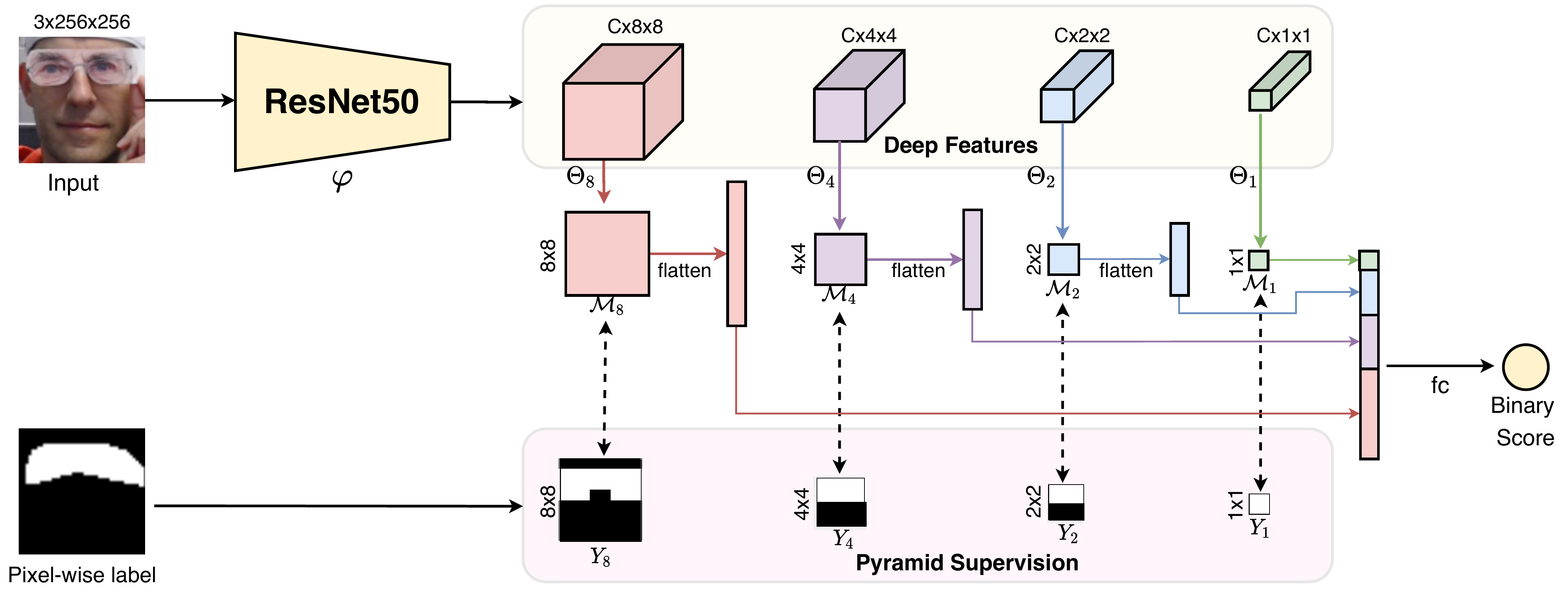}
  \caption{
  Pyramid supervision with multi-scale binary mask labels. In this framework, the deep features are firstly extracted from ResNet50, and downsampled into multi-scale representation. Unshared 1x1 convulotions cascaded with `Sigmoid' function are utilized for features mapping and predicting the multi-scale binary masks. Then multi-scale binary mask labels would provide sufficient pyramid supervision signals for training. Finally, the predicted multi-scale masks are flatted and concatenated for final live/spoof classification.
  }
\label{fig:PS1}
\end{figure*}

\noindent\textbf{Supervised by Others.}\quad     Besides the mainstream depth map and binary mask labels, there are several informative pixel-wise supervision (e.g., pseudo reflection map~\cite{kim2019basn,yu2020face,zhang2020celeba}, 3D point cloud map~\cite{li3dpc}, and ternary map~\cite{sun2020face}). According to the discrepancy of facial material-related albedo between the live skin and spoof mediums, Kim et al.~\cite{kim2019basn} proposed to supervise deep models with both depth and reflection labels. Moreover, Yu et al.~\cite{yu2020face} trained the bilateral convolutional networks with multiple pixel-wise supervisions (binary mask, reflection map, and depth map) simultaneously. Similarly, Zhang et al.~\cite{zhang2020celeba} adopted pseudo reflection and depth labels to guide the networks to learn rich semantic features. Unlike binary mask label considering all spatial positions, Sun et al.~\cite{sun2020face} removed the face-unrelated parts and left the entire face regions as a refined binary mask called `ternary map', which eliminated the noise outside the face. In order to reduce the redundancy from the dense depth map, Li et al.~\cite{li3dpc} used a sparse 3D point cloud map to efficiently supervise the lightweight models. In addition, encoder-decoder networks with the supervision from Fourier map~\cite{jourabloo2018face}, LBP texture map~\cite{zhang2020face}, zero map for live~\cite{feng2020learning}, and original RGB inputs~\cite{chen2019towards}, also showed their excellent detailed representation capacities.  

Although existing pixel-wise supervisions could provide fine-grained local information, there are still two limitations: 1) feature representation gap between pixel-wise supervision (local detailed features) and binary classification (semantic features); and 2) variant face resolution as well as PAs size among training/testing samples, which needs strict and exhausted pixel-wise annotated labels for quality assurance. In this paper, we propose the concept of pyramid supervision, which could be applied to existing pixel-wise supervision to alleviate the above-mentioned drawbacks.

\section{Methodology} \label{sec:pyramid}

As can be seen from Table~\ref{tab:review}, the mainstream backbones for pixel-wise supervision based FAS could be divided into two categories: 1) classical classification based networks (e.g., ResNet~\cite{He2015Deep} and DenseNet~\cite{huang2017densely}) with binary mask supervision; and 2) multi-scale fully convolutional networks (e.g., DepthNet~\cite{Liu2018Learning}) with pseudo depth supervision. In this section, we introduce the novel \textbf{pyramid supervision}, which is able to plug in these two categories flexibly. Here we adopt famous ResNet50~\cite{He2015Deep} and CDCN~\cite{yu2020searching} as the baselines for these two categories, respectively. The ablation studies of other architectures will be conducted in Section~\ref{sec:Ablation}.

\subsection{Pyramid Binary Mask Supervision}

Compared with binary scalar label, binary mask label~\cite{liu2019deep,george2019deep,hossaindeeppixbis,yu2020auto2,ma2020novel,liu2020disentangling,sun2020face2} contains rich spatial context, which benefits the models' cognition about the position of spoofing attacks. In order to exploit multi-scale spatial information from binary mask label, we propose the concept of pyramid supervision, which decomposes the original pixel-wise label into multiple spatial scales for supervising multi-scale features. There are three main advantages of pyramid supervision: 1) with observation and guidance of the pyramid labels, the models are able to learn features from multiple perspectives (from local details to global semantics) simultaneously; 2) the decision relies on multi-scale predicted results instead of any specific spatial level, which is more robust due to rich context evidence; and 3) the model could predict the multi-scale binary maps on-the-fly, which enhances the interpretability and localizes the spoofing attacks in different granularities.

As shown in Fig.~\ref{fig:PS1}, given a detected RGB face image with size 3$\times$256$\times$256 (Channel$\times$Height$\times$Width) as input $X$, the deep features $\mathcal{F}_{8}$ with size $C\times$8$\times$8 could be extracted via forwarding the backbone networks $\varphi$ (e.g., ResNet50). Then average pooling operators with different kernel size and strides are performed on $\mathcal{F}_{8}$ to generate multi-scale features $\mathcal{F}_{4}$, $\mathcal{F}_{2}$ and $\mathcal{F}_{1}$, respectively. Subsequently, respective feature-to-mask mappings ($\Theta_{8}$, $\Theta_{4}$, $\Theta_{2}$ and $\Theta_{1}$) with 1x1 convolution are conducted on multi-scale features ($\mathcal{F}_{8}$, $\mathcal{F}_{4}$, $\mathcal{F}_{2}$ and $\mathcal{F}_{1}$) to predict the corresponding binary masks ($\mathcal{M}_{8}$, $\mathcal{M}_{4}$, $\mathcal{M}_{2}$ and $\mathcal{M}_{1}$), which are then concatenated for final binary classification. The procedure of multi-scale binary mask prediction could be formulated as:
\begin{equation} 
\mathcal{M}_{i}=Sigmoid(\Theta_{i}(AvgPool_{i}(\varphi(X)))), i=\left\{ 8,4,2,1\right\},
\end{equation}
where the setting $kernel \& stride$=$1, 2, 4, 8$ is utilized for $AvgPool_{8}, AvgPool_{4}, AvgPool_{2}, AvgPool_{1}$, respectively. The $Sigmoid$ function guarantees the predicted masks ranging from [0,1].

\textbf{Loss Function.}\quad  In terms of the pixel-wise ground truth $Y$, we could use the already-annotated binary mask labels directly or the generated coarse binary masks. Then we decompose original mask labels into multi-scale mask labels $Y_{8}, Y_{4}, Y_{2}, and \; Y_{1}$ with the following criterion: as multi-scale labels are sub-sampled from $Y$, the positional values are determined by whether the corresponding local receptive fields have the spoofing attacks or not (see Fig.~\ref{fig:PS1} bottom for instance). As the predicted multi-scale masks and the ground truth have the same size, the pyramid loss $\mathcal{L}_{pyramid}$ can be calculated via accumulating the binary cross-entropy loss (BCE) from each position within each scale:
\begin{equation} 
\mathcal{L}_{pyramid}=\sum_{i=8,4,2,1}-(Y_{i}log(\mathcal{M}_{i})+(1-Y_{i})log(1-\mathcal{M}_{i})).
\end{equation}

In the training stage, the overall loss $\mathcal{L}_{overall}$ can be formulated as $\mathcal{L}_{overall}=\mathcal{L}_{pyramid}+\mathcal{L}_{binary}$, where the latter is the BCE loss for final classification. In the evaluation stage, for simplicity, only the final binary score is utilized for decision.

\textbf{Relation to Spatial Pyramid Pooling~\cite{he2015spatial}.}\quad  Here we discuss the relation between pyramid supervision and spatial pyramid pooling (SPP)~\cite{he2015spatial}, which share similar design philosophy but with different focuses. On one hand, SPP concatenates the deep multi-scale features directly for classification, which has huge feature dimension thus easily overfits in the FAS task (small-scale datasets). In contrast, in the framework of pyramid supervision, the predicted multi-scale compact maps are concatenated for final decision. On the other hand, SPP only provides the scalar-level supervision signal at the tail, while the proposed method could give multi-scale pixel-wise supervision before final classification. The ablation study is in Section 4.3 to show superior performance of pyramid supervision for FAS.

\begin{figure}[t]
\centering
\includegraphics[scale=0.36]{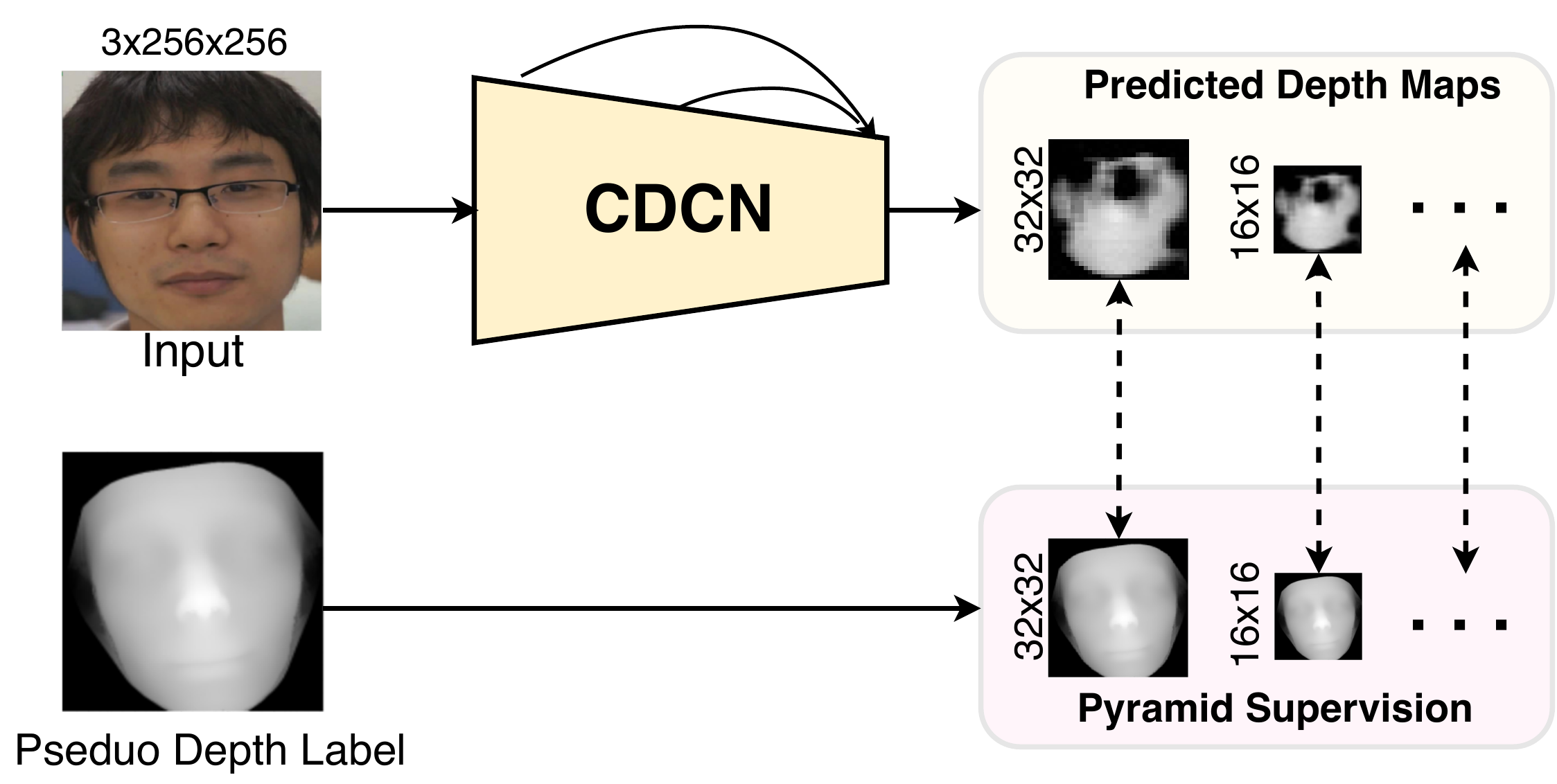}
  \caption{
  Pyramid supervision with multi-scale depth map labels. The pyramid depth loss could be calculated from the pair-wise errors between the predicted and ground truth depth maps with the same scale. 
  }
\label{fig:PS2}
\end{figure}

\subsection{Pyramid Depth Map Supervision}
Besides assembling deep models with pyramid binary mask supervision, we also show an example about applying pyramid depth map supervision in the recent CDCN. As shown in Fig.~\ref{fig:PS2}, CDCN extracts multi-level features from the face input with size 3×256×256, and predicts the grayscale facial depth with size 32×32. Similar to pyramid binary mask supervision, both the predicted depth map $\mathcal{D}_{32}$ and generated pseudo depth label $Y^{depth}$ are downsampled and resized to the same scales (32×32, 16×16, etc.). Here we adopt pyramid supervision with two scales (i.e., 32×32 and 16×16) as default setting, as we find that more scales make no contributions for performance improvement (see ablation study in Sec.\ref{sec:Ablation}). In terms of loss function, pyramid depth loss $\mathcal{L}^{depth}_{pyramid}$ can be formulated as
\begin{equation} 
\mathcal{L}^{depth}_{pyramid}=\sum_{i=32,16}(\mathcal{L}_{MSE}(\mathcal{D}_{i},Y^{depth}_{i})+\mathcal{L}_{CDL}(\mathcal{D}_{i},Y^{depth}_{i})),
\end{equation}
where $\mathcal{D}_{i}$ means the predicted depth map with scale $i$. $\mathcal{L}_{MSE}$ and $\mathcal{L}_{CDL}$ denotes mean square error loss and contrastive depth loss~\cite{wang2020deep}, respectively. In the training stage, only $\mathcal{L}^{depth}_{pyramid}$ is used for supervision. In the testing stage, we calculate the mean value of the predicted depth maps from all scales as the final score.

\section{Experiments}
\label{sec:experiment}

In this part, extensive experiments are conducted to demonstrate the effectiveness of the proposed pyramid supervision. In the following, we sequentially describe the employed datasets \& metrics (Sec. \ref{sec:dataset}), implementation details (Sec. \ref{sec:Details}), results (Sec. \ref{sec:Ablation} - \ref{sec:protocol2}) and visualization (Sec. \ref{sec:Analysis}).

\begin{figure}[t]
\centering
\includegraphics[scale=0.58]{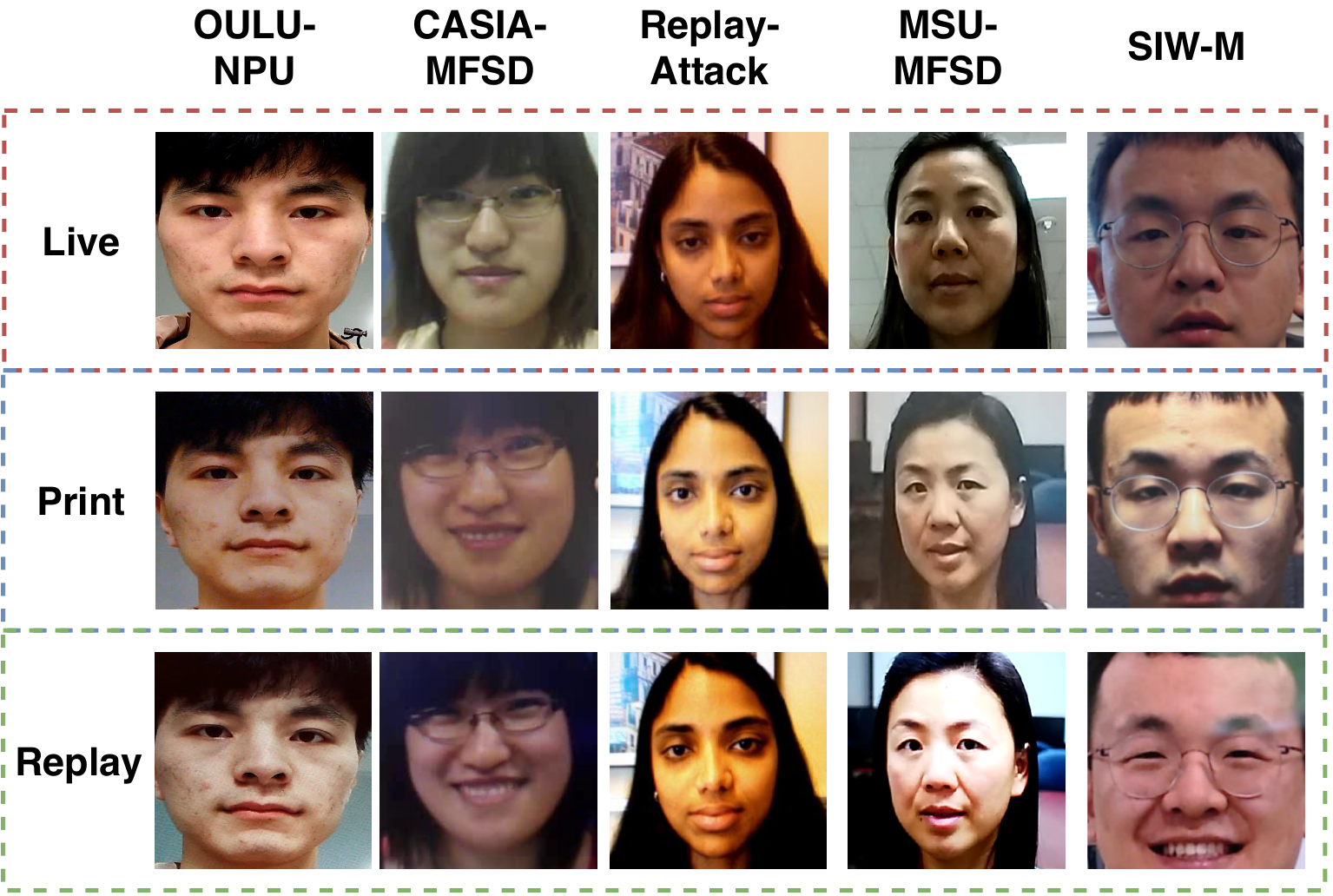}
  \caption{
  Visualization of the live and spoofing faces from the OULU-NPU~\cite{Boulkenafet2017OULU}, CASIA-MFSD~\cite{Zhang2012A}, Replay-Attack~\cite{ReplayAttack}, MSU-MFSD~\cite{wen2015face} and SiW-M~\cite{liu2019deep} datasets. It can be seen that serious domain shifts occur among these five datasets.
  }
\label{fig:dataset}
\end{figure}

\subsection{Databases and Metrics}
\label{sec:dataset}
\textbf{Datasets.}\quad
Five datasets including OULU-NPU~\cite{Boulkenafet2017OULU}, CASIA-MFSD~\cite{Zhang2012A}, Replay-Attack~\cite{ReplayAttack}, MSU-MFSD~\cite{wen2015face} and SiW-M~\cite{liu2019deep} are used in our experiments. OULU-NPU is a high-resolution database, containing four sub-protocols to validate the generalization (e.g., unseen illumination and attack medium) of models, which is used for intra-dataset testing. The videos in CASIA-MFSD, Replay Attack, MSU-MFSD and OULU-NPU are recorded under quite different scenarios with variant cameras and subjects, which are used for cross-dataset domain generalization protocol. As the SiW-M dataset has rich (13) attack types inside, we follow the original intra-dataset cross-type protocol on SIW-M to evaluate the generalization ability for unseen attack types.  Some typical bonafide and PAs samples from these five datasets are showed in Fig.~\ref{fig:dataset}.

\textbf{Evaluation Metrics.}\quad
In the OULU-NPU dataset, we follow the original sub-protocols and metrics, i.e., Attack Presentation Classification Error Rate (APCER), Bonafide Presentation Classification Error Rate (BPCER), and ACER~\cite{ACER} for a fair comparison. Area Under Curve (AUC) and Half Total Error Rate (HTER) are utilized for the cross-dataset domain generalization protocol on CASIA-MFSD, Replay-Attack, MSU-MFSD and OULU-NPU. For the intra-dataset cross-type protocol on SiW-M, 
both ACER and Equal Error Rate (EER) are employed.

\subsection{Implementation Details}
\label{sec:Details}

\textbf{Ground Truth Generation.}\quad
 The pseudo facial depth map label is generated by the off-the-shelf 3D face model~\cite{guo2020towards}. Apart from the SIW-M (already-annotated labels), the binary mask labels are generated simply by filling each position with corresponding binary label for other four datasets. The generated binary and depth maps are then downsampled into multiple resolution ($32\times32$, $16\times16$, $8\times8$, etc.) . The live depth map is normalized in a range of $[0, 1]$, while the spoof one is all 0 at the training stage, which is beneficial for learning discriminative patterns for the FAS task.

\textbf{Experimental Setting.}\quad 
Our proposed method is implemented with Pytorch. The ImageNet pretrained classification based models (e.g., ResNet and DenseNet) are trained on the FAS task using SGD optimizer with the initial learning rate (lr), momentum, and weight decay (wd) are 1e-3, 0.9, and 5e-5, respectively. We train models with maximum 80 epochs and batchsize 32 on single Nvidia V100 GPU. As for the multi-scale FCN models (e.g., CDCN with $\theta=0.7$ and DepthNet), Adam optimizer with initial lr=1e-4 and wd=5e-5 is used. The models are trained with batchsize 8 for maximum 800 epochs while lr halves at the 500th epoch.

\subsection{Ablation Study}
\label{sec:Ablation}
In this subsection, all ablation studies are conducted on the most challenging Protocol-4 (different illumination conditions, locations, cameras and attacks between training and testing sets) of OULU-NPU~\cite{Boulkenafet2017OULU} to explore the properties of pixel-wise supervision as well as the details of our proposed pyramid supervision.

\textbf{Impact of Pixel-Wise Supervision for Deep Models.}\quad 
Here we fully evaluate the performance of two models (ResNet50 and CDCN) with variant supervision signals, which are shown in Table~\ref{tab:ablation}. It can be seen from the 1st, 3rd and 4th rows that binary mask supervision sharply improves the performance of ResNet50 with either binary score or depth map supervision. It is surprising that supervision with depth map label performs the worst. This might be caused by the contradiction between semantic representation learning and the fine-grained depth guidance. One more interesting finding is that when replacing binary mask with multi-class mask labels (3 classes, i.e., live, print and replay are used in OULU-NPU), there is no performance gains achieved.
With more detailed spoofing cues, supervision with multi-class pixel-wise label is worth rethinking and exploring in the future. In terms of the CDCN, it is reasonable that supervision with depth map performs better than that with binary mask. In the following experiments, we adopt the configurations, i.e., ResNet50 with four-scale (8x8, 4x4, 2x2, 1x1) binary mask supervision and CDCN with two-scale (32x32, 16x16) depth map supervision.

\textbf{Impact of Scales of Pyramid Supervision.}\quad 
It can be seen from Table~\ref{tab:ablation} that, 1) pyramid binary mask supervision with four scales (8x8, 4x4, 2x2, 1x1) has the largest gain (-1.9$\pm$1.0\% ACER) for ResNet50 compared with the vanilla one scale (8x8 only) setting; and 2) pyramid depth map supervision with two scales (32x32, 16x16) significantly boosts the vanilla CDCN with only 32x32 depth supervision for -2.1$\pm$1.1\% ACER. As a result, these two kinds of scales are utilized as the default settings for pyramid binary mask and depth map supervision, respectively. It is interesting to find that more scales always benefit the ResNet with pyramid binary mask supervision but not the CDCN with pyramid depth map supervision. This is possibly because 1) pyramid binary mask supervision is likely to provide multi-scale fine-grained signals for ResNet50 to learn reasonable semantics; and 2) lower-resolution depth map labels might disturb CDCN to learn the local detailed representation.

\begin{table}[t]\small
\centering
\caption{The ablation study about scales of pyramid supervision with binary mask and depth map labels. `SPP' denotes Spatial Pyramid Pooling~\cite{he2015spatial}.}
\scalebox{0.88}{\begin{tabular}{|l|c | c|c|}
\hline
Model & Pixel-wise Label & Scale &ACER(\%)\\
\hline
ResNet50 & -  & -     & 7.1$\pm$3.4 \\
ResNet50 & - & 8x8, 4x4, 2x2, 1x1 (SPP)   & 8.9$\pm$4.5 \\
\hline
ResNet50 & depth map & 8x8   & 9.8$\pm$5.1 \\
ResNet50 & binary mask & 8x8   & 6.3$\pm$3.7 \\
ResNet50 & binary mask & 8x8, 4x4   & 5.4$\pm$3.0 \\
ResNet50 & binary mask & 4x4, 1x1   & 5.6$\pm$2.5 \\
ResNet50 & binary mask & 2x2, 1x1   & 5.8$\pm$1.5 \\
ResNet50 & binary mask & 8x8, 4x4, 2x2   & 4.8$\pm$2.8 \\
ResNet50 & binary mask & 8x8, 2x2, 1x1   & 4.4$\pm$3.0 \\
ResNet50 & binary mask & 4x4, 2x2, 1x1   & 5.2$\pm$2.4 \\
ResNet50 & binary mask & 8x8, 4x4, 2x2, 1x1   & \textbf{4.4$\pm$2.7} \\
\hline 

ResNet50 & multi-class mask & 8x8   & 6.7$\pm$2.5 \\
ResNet50 & multi-class mask & 8x8, 4x4, 2x2, 1x1   & 5.4$\pm$1.7 \\
\hline
\hline
CDCN & binary mask & 32x32   & 7.5$\pm$4.7 \\
CDCN & depth map & 32x32   & 6.9$\pm$2.9 \\
CDCN & depth map & 32x32, 16x16   & \textbf{4.8$\pm$1.8} \\
CDCN & depth map & 32x32, 16x16, 8x8   & 5.4$\pm$2.5 \\
CDCN & depth map & 32x32, 8x8, 4x4   & 6.7$\pm$3.7 \\
CDCN & depth map & 32x32, 16x16, 8x8, 4x4   & 7.5$\pm$2.9 \\
\hline

\end{tabular}}
\label{tab:ablation}
\vspace{-1.0em}
\end{table}

\begin{figure}
\centering
\includegraphics[scale=0.48]{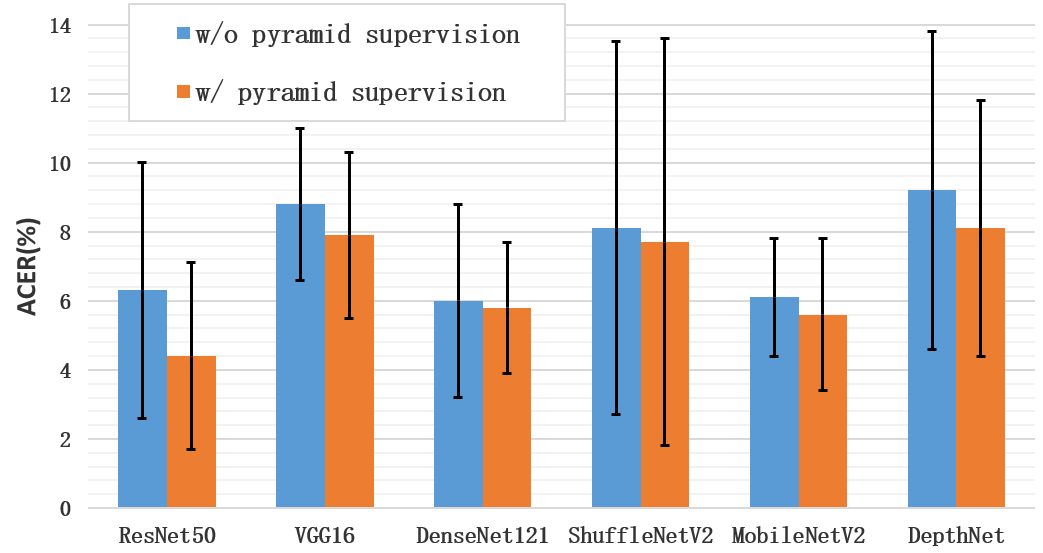}
  \vspace{-1.6em}
  \caption{
  Performance of various architectures w/ and w/o pyramid supervision on Protocol-4 of OULU-NPU. The pyramid supervision for `DepthNet' is with two-scale (32x32, 16x16) depth maps while the other architectures are with four-scale (8x8, 4x4, 2x2, 1x1) binary masks.}
 
\label{fig:arhitectures}
\end{figure}

\textbf{Pyramid Supervision vs. Spatial Pyramid Pooling (SPP)~\cite{he2015spatial}.}\quad 
As pyramid supervision and SPP share the similar insight of multi-scale features, here we also study the performance of SPP. Specifically, SPP module is cascaded with the deep features from ResNet50, and then binary score label is utilized for supervision. Despite with gained performance in generic object detection and classification tasks, SPP could not benefit the FAS task, and even performs 1.8$\pm$1.1\% worse ACER than the counterpart without SPP. It is possibly caused by the overfitting from SPP module with high-dimensional features. In contrast, the proposed pyramid binary mask supervision with scales `8x8, 4x4, 2x2, 1x1' outperforms ResNet50 with SPP by a large margin (4.5$\pm$1.8\% ACER), indicating the efficiency of the pyramid supervision for the predicted compact multi-scale binary masks.

\textbf{Generalization to Architectures.}\quad
Besides ResNet50 and CDCN, we also study the generalization ability on more architectures. As illustrated in Fig.~\ref{fig:arhitectures}, pyramid binary mask supervision improves both large models (VGG16~\cite{simonyan2014very} and DenseNet121~\cite{huang2017densely}) and mobile models (ShuffleNetV2~\cite{ma2018shufflenet} and MobileNetV2~\cite{sandler2018mobilenetv2}) consistently, indicating the promising application of the proposed pyramid supervision for further vision models. In addition, with pyramid depth map supervision, both the mean and standard deviation ACER are decreased on DepthNet, which shows the excellent generalization of pyramid supervision.

\begin{table}[t]
\centering
\caption{The results of intra testing on OULU-NPU~\cite{Boulkenafet2017OULU}. `\_8x8' and `-PS' denote with 8x8 binary mask and pyramid supervision, respectively.}
\resizebox{0.49\textwidth}{!}{
\begin{tabular}{|c|c|c|c|c|}

\hline
Prot. & Method & APCER(\%)$\downarrow$ & BPCER(\%)$\downarrow$ & ACER(\%)$\downarrow$ \\
\hline
\multirow{6}{*}{1}
        &STASN ~\cite{yang2019face} &1.2 &2.5 & 1.9 \\
        &Auxiliary ~\cite{Liu2018Learning} &1.6 &1.6 & 1.6 \\
        &FaceDs ~\cite{jourabloo2018face} &1.2 &1.7 & 1.5 \\
        &FAS-SGTD ~\cite{wang2020deep} &2.0 &0.0 & 1.0 \\
        &SpoofTrace ~\cite{liu2020disentangling} &0.8 &1.3 & 1.1 \\
        &Disentangled ~\cite{zhang2020face} &1.7 &0.8 & 1.3 \\
        &BCN ~\cite{yu2020face} &0.0 &1.6 & 0.8 \\
        &DeepPixBiS ~\cite{george2019deep}&0.8 &0.0 & \textbf{0.4} \\
        \cline{2-5}
        &ResNet50\_8x8 ~\cite{He2015Deep}& 2.3
       &6.7 &4.5 \\
       &\textbf{ResNet50-PS (Ours)} &1.9 &6.4  &4.1 \\
       &CDCN ~\cite{yu2020searching}&0.4 &1.7 & 1.0 \\
       &\textbf{CDCN-PS (Ours)} &0.4 &1.2 &0.8 \\
\hline
\hline
\multirow{6}{*}{2} 
       &DeepPixBiS ~\cite{george2019deep}&11.4 &0.6 & 6.0 \\
       &FaceDs ~\cite{jourabloo2018face}&4.2 &4.4 & 4.3 \\
       &Auxiliary ~\cite{Liu2018Learning}&2.7 &2.7 & 2.7 \\
       &STASN ~\cite{yang2019face}&4.2 &0.3 & 2.2 \\
       &FAS-SGTD ~\cite{wang2020deep} &2.5 & 1.3 & 1.9 \\
       &Disentangled ~\cite{zhang2020face} &1.1 &3.6 & 2.4 \\
       &BCN ~\cite{yu2020face} &2.6 &0.8 & 1.7 \\
       &SpoofTrace ~\cite{liu2020disentangling} &2.3 &1.6 & 1.9 \\
       \cline{2-5}
       &ResNet50\_8x8 ~\cite{He2015Deep}& 0.6
       &10 &5.3 \\
       &\textbf{ResNet50-PS (Ours)} &1.9 &7.5  &4.7 \\
       &CDCN ~\cite{yu2020searching}&1.5 &1.4 & 1.5 \\
       &\textbf{CDCN-PS (Ours)} &1.4 &1.4 &\textbf{1.4} \\
\hline
\hline
\multirow{4}{*}{3} 
       &DeepPixBiS ~\cite{george2019deep}&11.7$\pm$19.6 &10.6$\pm$14.1 & 11.1$\pm$9.4 \\
       &FAS-SGTD ~\cite{wang2020deep}&3.2$\pm$2.0 & 2.2$\pm$1.4 & 2.7$\pm$0.6 \\
       &FaceDs ~\cite{jourabloo2018face}&4.0$\pm$1.8 &3.8$\pm$1.2 &3.6$\pm$1.6 \\
       &Auxiliary ~\cite{Liu2018Learning}&2.7$\pm$1.3 &3.1$\pm$1.7 &{2.9}$\pm$1.5 \\
       &STASN ~\cite{yang2019face}&4.7$\pm$3.9 &0.9$\pm$1.2  &2.8$\pm$1.6 \\
       &Disentangled ~\cite{zhang2020face}&2.8$\pm$2.2 &1.7$\pm$2.6  &2.2$\pm$2.2 \\
       &SpoofTrace ~\cite{liu2020disentangling}&1.6 $\pm$1.6 & 4.0$\pm$5.4  &2.8$\pm$3.3 \\
       &BCN ~\cite{yu2020face}&2.8$\pm$2.4 &2.3$\pm$2.8  &2.5$\pm$1.1 \\
       \cline{2-5}
       &ResNet50\_8x8 ~\cite{He2015Deep}& 2.9$\pm$1.2
       &4.2$\pm$1.9 &3.6$\pm$1.4 \\
       &\textbf{ResNet50-PS (Ours)} &2.2$\pm$1.4 &4.0$\pm$1.7  &3.1$\pm$1.6 \\
       &CDCN ~\cite{yu2020searching}&2.4$\pm$1.3 &2.2$\pm$2.0  &2.3$\pm$1.4 \\
       &\textbf{CDCN-PS (Ours)} &1.9$\pm$1.7 &2.0$\pm$1.8  &\textbf{2.0$\pm$1.7} \\

\hline
\hline
\multirow{4}{*}{4} 
        &DeepPixBiS ~\cite{george2019deep}&36.7$\pm$29.7 &13.3$\pm$14.1 & 25.0$\pm$12.7 \\
       &Auxiliary ~\cite{Liu2018Learning}&9.3$\pm$5.6 &10.4$\pm$6.0 &9.5$\pm$6.0 \\
       &FAS-SGTD ~\cite{wang2020deep}&6.7$\pm$7.5 &3.3$\pm$4.1 & 5.0$\pm$2.2 \\
       &STASN ~\cite{yang2019face}&6.7$\pm$10.6 &8.3$\pm$8.4  &7.5$\pm$4.7 \\
       &FaceDs ~\cite{jourabloo2018face}&1.2$\pm$6.3
       &6.1$\pm$5.1 &5.6$\pm$5.7 \\
       &Disentangled ~\cite{zhang2020face}&5.4$\pm$2.9 &3.3$\pm$6.0  &4.4$\pm$3.0 \\
       &SpoofTrace ~\cite{liu2020disentangling}&2.3$\pm$3.6 &5.2$\pm$5.4  &\textbf{3.8$\pm$4.2} \\
       &BCN ~\cite{yu2020face}&2.9$\pm$4.0 &7.5$\pm$6.9  &5.2$\pm$3.7 \\
       \cline{2-5}
       &ResNet50\_8x8 ~\cite{He2015Deep}& 5.8$\pm$3.2
       &6.8$\pm$4.2 &6.3$\pm$3.7 \\
       &\textbf{ResNet50-PS (Ours)} &2.9$\pm$4.0 &5.8$\pm$4.9  &4.4$\pm$2.7 \\
       &CDCN ~\cite{yu2020searching}&4.6$\pm$4.6 &9.2$\pm$8.0  &6.9$\pm$2.9 \\
       &\textbf{CDCN-PS (Ours)} &2.9$\pm$4.0 &5.8$\pm$4.9  &4.8$\pm$1.8 \\
\hline
\end{tabular}
}
\label{tab:OULU}
\end{table}

\subsection{Intra-Dataset Intra-Type Testing on OULU-NPU}
\label{sec:protocol1}


We strictly follow the four protocols on OULU-NPU for the evaluation. As shown in Table~\ref{tab:OULU}, with our proposed pyramid supervision, the ACER are decreased on four protocols consistently for both ResNet50 (-0.4\%, -0.6\%, -0.5\%, and -1.9\%) and CDCN (-0.2\%, -0.1\%, -0.2\%, and -2.1\%), respectively. It indicates that pyramid supervision performs well at the generalization of the external environment, attack mediums and input camera variation. 

In terms of architectures, `CDCN-PS' performs better than or on par with the state-of-the-art methods in all four protocols. As for `ResNet50-PS', despite with bad performance for the first three protocols, it surprisingly achieves better performance than `CDCN-PS' in the most challenging Protocol-4, which indicates the efficacy of pyramid binary mask supervision even with limited training samples. In the future, it is promising to plug pyramid supervision into more advanced architectures as well as pixel-wise labels.

\begin{table*}
\centering
\caption{Results of cross-dataset intra-type testing on OULU-NPU, CASIA-MFSD, Replay-Attack, and MSU-MFSD. }

\scalebox{1.0}{\begin{tabular}{c|c|c|c|c|c|c|c|c}
\hline
\multirow{2}{*}{\textbf{Method}} &\multicolumn{2}{c|}{\textbf{O\&C\&I to M}} &\multicolumn{2}{c|}{\textbf{O\&M\&I to C}}&\multicolumn{2}{c|}{\textbf{O\&C\&M to I}} &\multicolumn{2}{c}{\textbf{I\&C\&M to O}} \\
\cline{2-9} &\tabincell{c}{HTER(\%)$\downarrow$} &\tabincell{c}{AUC(\%)$\uparrow$} &\tabincell{c}{HTER(\%)$\downarrow$} &\tabincell{c}{AUC(\%)$\uparrow$}&\tabincell{c}{HTER(\%)$\downarrow$}&\tabincell{c}{AUC(\%)$\uparrow$}&\tabincell{c}{HTER(\%)$\downarrow$}&\tabincell{c}{AUC(\%)$\uparrow$} \\
\hline
Color Texture ~\cite{boulkenafet2016face} 
& 28.09 & 78.47 & 30.58 & 76.89 & 40.40 & 62.78 & 63.59 & 32.71 \\

Auxiliary (Depth) ~\cite{Liu2018Learning} 
& 22.72 & 85.88 & 33.52 & 73.15 & 29.14 & 71.69 & 30.17 & 66.61 \\

MMD-AAE ~\cite{li2018domain} 
& 27.08 & 83.19 & 44.59 & 58.29 & 31.58 & 75.18 & 40.98 & 63.08 \\

MADDG ~\cite{shao2019multi} 
& 17.69 & 88.06 & 24.50 & 84.51 & 22.19 & 84.99 & 27.98 & 80.02 \\

DR-MD-Net ~\cite{wang2020cross} 
& 17.02 & 90.10 & 19.68 & 87.43 & 20.87 & 86.72 & 25.02 & 81.47 \\

OCA-FAS ~\cite{qin2020one} 
& 15.51 & 91.92 & 18.45 & \underline{89.33} & 19.51 & 88.29 & 19.25 & 88.55 \\

RFMeta ~\cite{shao2019regularized} 
& \textbf{13.89} & \underline{93.98} & 20.27 & 88.16 & \textbf{17.30} & \textbf{90.48} & \underline{16.45} & \underline{91.16} \\
\hline
ResNet50\_8x8 ~\cite{He2015Deep}
& 18.80 & 90.48 & \underline{18.26} & 86.42 & 19.83 & 88.24 & 26.47 & 80.72\\

\textbf{ResNet50-PS (Ours)}
& \underline{14.32} & \textbf{94.51} & \textbf{18.23} & \textbf{89.75} & \underline{18.86} & \underline{89.63} & 21.44 & 87.56\\

CDCN ~\cite{yu2020searching}
& 22.90 & 85.45 & 22.46 & 86.64 & 19.98 & 84.75 & 16.92 & 90.46\\

\textbf{CDCN-PS (Ours)}
& 20.42 & 87.43 & 18.25 & 86.76 & 19.55 & 86.38 & \textbf{15.76} & \textbf{92.43}\\
\hline

\end{tabular}}

\label{tab:DG}
\end{table*}

\begin{figure}
\centering
\includegraphics[scale=0.5]{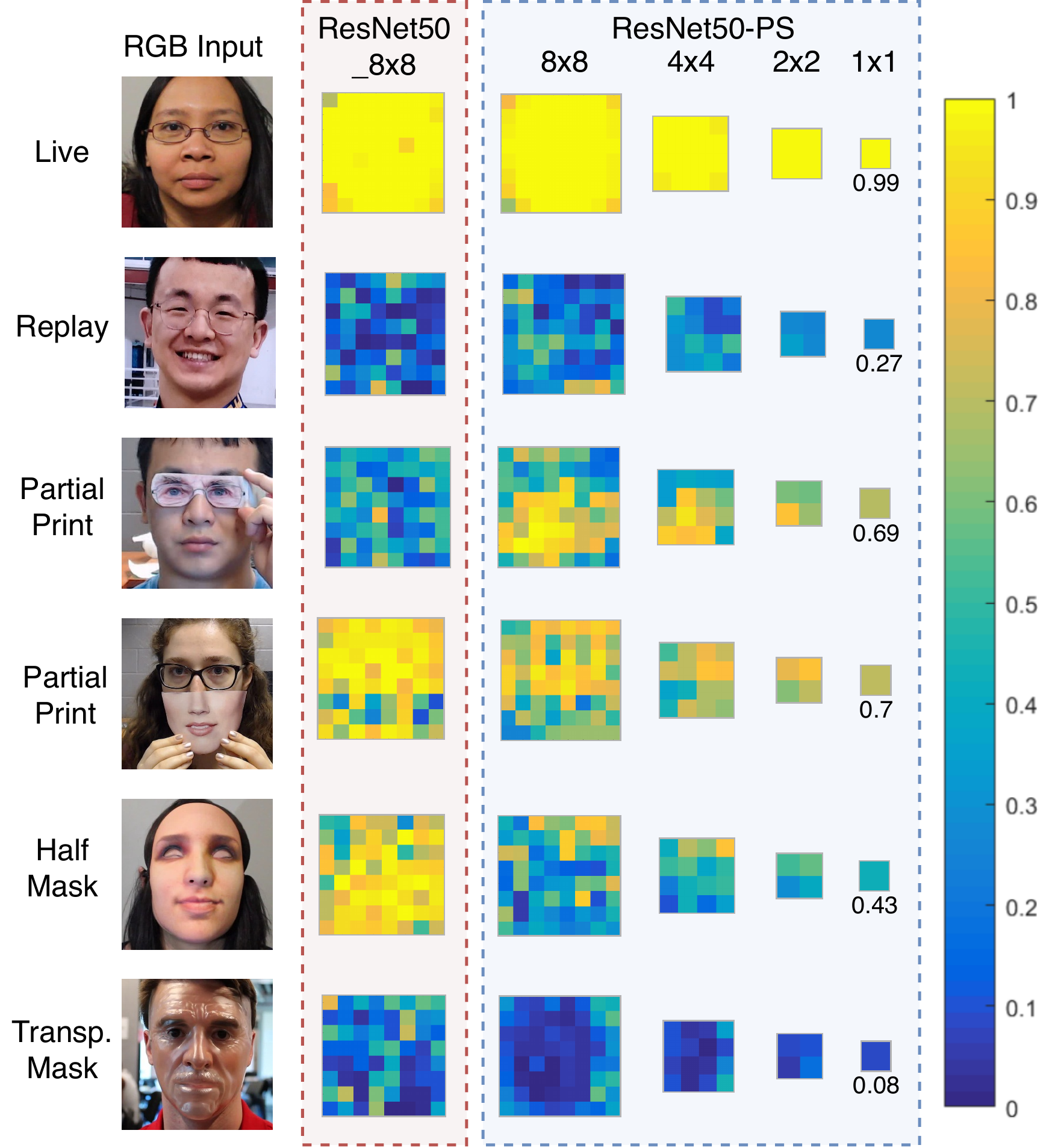}
  \caption{
  Visualization of the predicted binary masks from `ResNet50' and `ResNet50-PS' on SiW-M. Higher predicted scores (yellow) indicates higher probability of liveness. }
 
\label{fig:visualization}
\end{figure}

\subsection{Cross-Dataset Intra-Type Testing}
\label{sec:protocol3}

Four datasets OULU-NPU (O), CASIA-MFSD (C), Idiap Replay-Attack (I) and MSU-MFSD (M) are utilized here. Specifically, three datasets are randomly selected for training and the remained one leaves for testing. It can be seen from Table~\ref{tab:DG} that with pyramid binary mask supervision, the performances of `ResNet50-PS' are improved remarkably by -4.48\% and -5.03\% HTER on protocols `O\&C\&I to M’ and `I\&C\&M to O’, respectively. Similarly, CDCN with pyramid depth map supervision outperforms the vanilla CDCN by -2.48\%, -4.21\% and -1.16\% HTER on protocols `O\&C\&I to M’, `O\&M\&I to C’ and `I\&C\&M to O’, respectively. The experimental results indicate that the proposed pyramid supervision is also helpful for providing rich multi-scale guidance on multiple source domains.

\begin{table*}
\centering
\caption{Results of the cross-type testing on SiW-M~\cite{liu2019deep}.}

\scalebox{0.83}{\begin{tabular}{c|c|c|c|c|c|c|c|c|c|c|c|c|c|c|c}
\hline
\multirow{2}{*}{Method} &\multirow{2}{*}{Metrics(\%)} &\multirow{2}{*}{Replay} &\multirow{2}{*}{Print} &\multicolumn{5}{c|}{Mask Attacks} &\multicolumn{3}{c|}{Makeup Attacks}&\multicolumn{3}{c|}{Partial Attacks} &\multirow{2}{*}{Average} \\
\cline{5-15} &  &  &  & \tabincell{c}{Half} &\tabincell{c}{Silicone} &\tabincell{c}{Trans.} &\tabincell{c}{Paper}&\tabincell{c}{Manne.}&\tabincell{c}{Obfusc.}&\tabincell{c}{Im.}&\tabincell{c}{Cos.}&\tabincell{c}{Fun.} & \tabincell{c}{Glasses} &\tabincell{c}{Partial} & \\

\hline

\multirow{2}{*}{SVM+LBP~\cite{Boulkenafet2017OULU}} & ACER & 20.6 & 18.4 & 31.3 & 21.4 & 45.5 & 11.6 & 13.8 & 59.3 & 23.9 & 16.7 & 35.9 & 39.2 & 11.7 & 26.9$\pm$14.5 \\
  & EER & 20.8 & 18.6 & 36.3  & 21.4 & 37.2 & 7.5 & 14.1 & 51.2 & 19.8 & 16.1 & 34.4 & 33.0 & 7.9 & 24.5$\pm$12.9 \\

\hline

\multirow{2}{*}{Auxiliary~\cite{Liu2018Learning}} &  ACER & 16.8 & 6.9 & 19.3 & 14.9 & 52.1 & 8.0 & 12.8 & 55.8 & 13.7 & \textbf{11.7} & 49.0 & 40.5 & 5.3 & 23.6$\pm$18.5 \\
  & EER & 14.0 & 4.3 & 11.6  & 12.4 & 24.6 & 7.8 & 10.0 & 72.3 & 10.1 & \textbf{9.4} & 21.4 & 18.6 & 4.0 & 17.0$\pm$17.7 \\

\hline

\multirow{2}{*}{DTN~\cite{liu2019deep}}  & ACER & 9.8 & 6.0 & 15.0 & 18.7 & 36.0 & 4.5 & 7.7 & 48.1 & 11.4 & 14.2 & \textbf{19.3} & 19.8 & 8.5 & 16.8 $\pm$11.1 \\
 & EER & 10.0 & \textbf{2.1} & 14.4 & 18.6 & 26.5 & \textbf{5.7} & 9.6 & 50.2 & 10.1 & 13.2 & \textbf{19.8} & 20.5 & 8.8 & 16.1$\pm$ 12.2 \\

\hline

\multirow{2}{*}{SpoofTrace ~\cite{liu2020disentangling}}  & ACER & \textbf{7.8} & 7.3 & \textbf{7.1} & 12.9 & 13.9 & 4.3 & 6.7 & 53.2 & 4.6 & 19.5 & 20.7 & 21.0 & 5.6 & 14.2 $\pm$13.2\\
 & EER & \textbf{7.6} & 3.8 & 8.4 & 13.8 & 14.5 & 5.3 & 4.4 & 35.4 & \textbf{0.0} & 19.3 & 21.0 & 20.8 & 1.6 & 12.0$\pm$ 10.0 \\

\hline

\multirow{2}{*}{BCN ~\cite{yu2020face}}  & ACER & 12.8 & \textbf{5.7} & 10.7 & 10.3 & 14.9 & 1.9 & \textbf{2.4} & 32.3 & 0.8 & 12.9 & 22.9 & \textbf{16.5} & \textbf{1.7} & \textbf{11.2 $\pm$9.2}\\
 & EER & 13.4 & 5.2 & \textbf{8.3} & 9.7 & 13.6 & 5.8 & \textbf{2.5} & 33.8 & \textbf{0.0} & 14.0 & 23.3 & 16.6 & 1.2 & \textbf{11.3$\pm$ 9.5} \\

\hline
\hline

\multirow{2}{*}{ResNet50\_8x8 ~\cite{He2015Deep}} & ACER & 12.8 & 12.4 & 19.5 & 17.1 & 13.6 & 4.3 & 9.2 & 26.2 & 2.8 & 18.7 & 22.5 & 25.1 & 12.4 & 15.1 $\pm$7.3\\
 & EER & 12.4 & 10.4 & 18.1 & 18.5 & 10.2 & \textbf{0.0} & 7.5 & 26.1 & 1.6 & 18.0 & 20.1 & 24.6 & 11.6 & 13.8$\pm$ 8.0 \\

\hline

\multirow{2}{*}{\textbf{ResNet50-PS (Ours)}} & ACER & 10.6 & 10.6 & 20.2 & 15.9 & \textbf{10.0} & \textbf{1.2} & 10.0 & \textbf{23.3} & 0.4 & 14.0 & 20.5 & 24.3 & 5.8 & 12.8 $\pm$7.8\\
 & EER & 11.3 & 9.6 & 19.4 & 14.8 & \textbf{6.8} & \textbf{0.0} & 7.5 & \textbf{21.7} & \textbf{0.0} & 12.4 & 19.5 & 21.4 & 5.0 & 11.5$\pm$ 7.6 \\

\hline

\multirow{2}{*}{CDCN ~\cite{yu2020searching}} & ACER & 8.7 & 7.7 & 11.1 & \textbf{9.1} & 20.7 & 4.5 & 5.9 & 44.2 & 2.0 & 15.1 & 25.4 & 19.6 & 3.3 & 13.6 $\pm$11.7 \\
 & EER & 8.2 & 7.8 & \textbf{8.3} & \textbf{7.4} & 20.5 & 5.9 & 5.0 & 47.8 & 1.6 & 14.0 & 24.5 & 18.3 & 1.1 & 13.1$\pm$ 12.6 \\

\hline

\multirow{2}{*}{\textbf{CDCN-PS (Ours)}}  & ACER & 12.1 & 7.4 & 9.9 & \textbf{9.1} & 14.8 & 5.3 & 5.9 & 43.1 & \textbf{0.4} & 13.8 & 24.4 & 18.1 & 3.5 & 12.9$\pm$11.1 \\
  & EER & 10.3 & 7.8 & \textbf{8.3}  & \textbf{7.4} & 10.2 & 5.9 & 5.0 & 43.4 & \textbf{0.0} & 12.0 & 23.9 & \textbf{15.9} & \textbf{0.0} & 11.5$\pm$11.4 \\

\hline

\end{tabular}
}
\label{tab:SiW-M}
\end{table*}

\subsection{Intra-Dataset Cross-Type Testing on SiW-M}
\label{sec:protocol2}

Following the same cross-type testing protocol (13 attacks leave-one-out) on SiW-M, we compare our proposed methods with several recent FAS methods~\cite{Boulkenafet2017OULU,Liu2018Learning,liu2019deep,yu2020face,liu2020disentangling} to validate the generalization capacity of unseen attacks. As shown in Table~\ref{tab:SiW-M}, compared with the vanilla pixel-wise supervision, `ResNet50-PS' and `CDCN-PS' achieve an overall better EER with the improvement by 17\% and 12\% respectively. Specifically, benefitted from the pyramid supervision, our method can perform more robustly in the two challenging attacks (`Transparent Mask' and `Partial Paper') while the vanilla supervision performs poorly on these two attacks. Note that the proposed methods only with one specific pixel-wise label achieve comparable performance with the state-of-the-art method BCN~\cite{yu2020face} supervised by three kinds of pixel-wise labels (i.e., binary mask, depth map and reflection map). In the future, extending pyramid supervision on multiple pixel-wise labels simultaneously for exploiting richer and more fine-grained cues is favorable.


\subsection{Visualization and Analysis}
\label{sec:Analysis}

Fig.~\ref{fig:visualization} illustrates the predicted binary maps of live and spoof faces on the SiW-M dataset under cross-type testing protocol. On one hand, in terms of the predictions from `Live', `Replay' and `Transparent Mask', both `ResNet50\_8x8' and `ResNet50-PS' perform well and keep high confidence for discrimination. On the other hand, when fed with the unseen and challenging attack types (e.g., `Partial Print' and `Half Mask'), the predictions become less reliable and chaotic. To be specific, it can be seen from the `ResNet50\_8x8' results in 3rd row, 2nd column that besides the print parts in the eye regions, other facial regions are predicted as low liveness. The similar result also occurs in 3rd row, 2nd column with high liveness confidence in the mask region. In contrast, assembled with pyramid supervision, the interpretability about the spoof localization improves significantly. From the predicted 8x8 and 4x4 maps We can find that the high scores (liveness) and low activations (spoofing) position in the facial skin regions and spoof mediums, respectively. With the evolution of spoofing attacks, we believe that network interpretability becomes more and more important in spoof localization and understanding.

\section{Conclusion} \label{sec:conclusion}
In this paper, we give a elaborate review about face anti-spoofing (FAS) with pixel-wise supervision and conduct extensive experiments to study corresponding factors. Moreover, we propose a novel pyramid supervision,intending to provide richer multi-scale spatial context for fine-grained supervision. Experimental results demonstrate the effectiveness of our proposed pyramid supervision on both performance improvement and interpretability enhancement.

The possible future directions include: 1) discovering the optimal pixel-wise labels with suitable supervision strategies (e.g., pyramid supervision) automatically is worth exploring; and 2) large-scale FAS benchmark with fine-grained (i.e., multiple classes) pixel/patch-level annotated labels should be established, which also helps to quantitively evaluate the interpretability.

\section*{Acknowledgments} This work was supported by the Academy of Finland for project MiGA (grant 316765), ICT 2023 project (grant 328115), Infotech Oulu. As well, the authors wish to acknowledge CSC-IT Center for Science, Finland, for computational resources.

\bibliographystyle{IEEEtran}
\bibliography{IEEEabrv,reference}

\end{document}